\documentclass{article}

\usepackage[utf8]{inputenc} 
\usepackage[T1]{fontenc}    
\usepackage{hyperref}       
\usepackage{url}            
\usepackage{booktabs}       
\usepackage{amsfonts}       
\usepackage{nicefrac}       
\usepackage{microtype}      

\usepackage{graphicx} 
\usepackage{array}
\usepackage{amsmath}
\usepackage{placeins}
\usepackage{subcaption}
\usepackage{caption}
\usepackage{natbib}
\usepackage{fancyhdr}
\usepackage{tabularx}

\usepackage[
  a4paper,
  textheight=9in,
  textwidth=6in,
  top=1in
  ]{geometry}
  
\renewenvironment{abstract}%
{%
  \vskip 0.075in%
  \centerline%
  {\large\bf Abstract}%
  \vspace{0.5ex}%
  \begin{quote}%
}
{
  \par%
  \end{quote}%
  \vskip 1ex%
}

\date{}

\title{Recurrent Neural Networks With Limited Numerical Precision}

\author{
  Joachim Ott$^*$, Zhouhan Lin$^\ddagger$,Ying Zhang$^\ddagger$, Shih-Chii Liu$^*$, Yoshua Bengio$^\ddagger$$^\dag$ \\
  $^*$Institute of Neuroinformatics, University of Zurich and ETH Zurich \\
  \texttt{ottj@ethz.ch, shih@ini.ethz.ch} \\
  $^\ddagger$D\'epartement d'informatique et de recherche op\'erationnelle, Universit\'e de Montr\'eal \\
  $^\dag$CIFAR Senior Fellow \\
  \texttt{\{zhouhan.lin, ying.zhang\}@umontreal.ca} \\
}

\begin{document}
\maketitle

\begin{abstract}

Recurrent Neural Networks (RNNs) 
produce state-of-art performance on many machine learning tasks but 
their demand on resources in terms of memory and computational power are often
high. Therefore, there is a great interest in optimizing the computations performed
with these models especially when considering development of specialized
low-power hardware for deep networks. One way of reducing the computational needs is to limit the
numerical precision of the network weights and biases. This
has led to different proposed rounding methods which have been applied so far to only
Convolutional Neural Networks and Fully-Connected Networks.
This paper addresses the question of how to best reduce weight precision during training in the case of RNNs. We present results from
the use of different stochastic and deterministic reduced precision training methods applied to three major RNN types which are then tested
on several datasets.
The results show that the weight binarization methods do not work with
the RNNs. However, the stochastic and
deterministic ternarization, and pow2-ternarization methods gave rise to low-precision RNNs that produce similar
and even higher accuracy on certain datasets therefore providing a path towards training more efficient implementations of RNNs
in specialized hardware.

\end{abstract}

\vspace*{-2mm}
\section{Introduction}
\vspace*{-1mm}

A Recurrent Neural Network (RNN) is a specific type of neural network which is able to process
input and output sequences of variable length. Because of this nature, RNNs are suitable for sequence modeling. Various RNN architectures have been proposed in recent years, based on different forms of non-linearity, such as the Gated Recurrent Unit (GRU) \citep{Cho2014} and Long-Short Term Memory (LSTM) \citep{Hochreiter1997}. They have enabled new levels of performance in many tasks such as speech recognition \citep{Amodei2015}\citep{Chan2015}, machine translation \citep{Devlin2014}\citep{Chung2016}\citep{Sutskever2014}, or even video games \citep{Mnih2015} and Go\citep{Silver2016}.

Compared to standard feed-forward networks, RNNs often take longer to train and are more demanding in memory and computational power. For example, it can take weeks to train models for state-of-the-art machine translation and speech recognition. Thus it is of vital importance to accelerate computation and reduce training time of such networks. On the other hand, even at run-time, these models require too much in terms of computational resources if we want to deploy a model onto low-power embedded hardware devices. Increasingly, dedicated deep learning hardware platforms including FPGAs~\citep{Farabet_etal11} and custom chips~\citep{Sim_etal2016} are reporting higher computational efficiencies of up to tera operations per second per watt (TOPS/W). These platforms are targeted at deep CNNs. If low-precision RNNs are able to report the same performance, then the savings in the reduction of multipliers (the circuits that take the space and energy) and memory storage of the weights would be even larger as the bit precision of the multipliers needed for the 2 to 3 gates of the gated RNN units can be reduced or the multipliers removed completely.

Previous work showed the successful application of stochastic rounding strategies on feed forward networks, including binarization \citep{Courbariaux2015} and ternarization \citep{Lin2015} of weights of vanilla Deep Neural Networks (DNNs) and Convolutional Neural Networks (CNNs) \citep{Rastegari2016}, and in \citep{Courbariaux2016} even the quantization of their activations, during training and run-time. Quantization of RNN weights has so far only been used with pretrained models \cite{shin2016fixed}.  

What remained an open question up to now was whether these weight quantization techniques could successfully be applied to RNNs during training.

In this paper, we use different methods to reduce the numerical precision of
weights in RNNs, and test their performance on different benchmark
datasets. We make the code for the rounding methods available. \footnote{\url{https://github.com/ottj/QuantizedRNN}}
We use three popular RNN models: vanilla RNNs, GRUs, and LSTMs. Section
\ref{lowprecision} covers the 4 ways of obtaining low-precision weights
for the RNN models in this work, and Section \ref{models} elaborates on the test results of the 
low-precision RNN models on different datasets including the large WSJ dataset.
We find that ternary quantization works very well while binary quantization fails and we analyze this result.

\vspace*{-2mm}

\section{Rounding Network Weights}  
\vspace*{-1mm}
\label{lowprecision}

This work evaluates the use of 4 different rounding methods on the weights of
various types of RNNs. These methods include the stochastic and 
deterministic binarization method (BinaryConnect) \citep{Courbariaux2016} and ternarization method (TernaryConnect)
\citep{Lin2015}, the pow2-ternarization method \citep{Stromatias2015},
and a new weight quantization method (Section
\ref{quantization}). For all 4 methods, we keep a full-precision copy of the
weights and biases during training to accumulate the small updates, while during test time, we
can either use the learned full-precision weights or use their deterministic
low-precision version. As experimental results in Section \ref{exp} show, the
network with learned full-precision weights usually yields better results than
a baseline network with full precision during training, due to the extra
regularization effect brought by stochastic binarization. The deterministic
low-precision version could still yield comparable performance while
drastically reducing computation and required memory storage at test time. We
will briefly describe the former 3 low-precision methods, and introduce a new
fourth method called Exponential Quantization.

\vspace*{-1mm}
\subsection{Binarization, Ternarization, and Pow2-Ternarization}
\vspace*{-1mm}
BinaryConnect and TernaryConnect were first introduced in
\citep{Courbariaux2015} and \citep{Lin2015} respectively. By limiting the weights
to only 2 or 3 possible values, i.e., -1 or 1 for BinaryConnect and -1, 0 or 1
for TernaryConnect, these methods do not require the use of multiplications. In
the stochastic versions of both methods, the low-precision weights are obtained
by stochastic sampling, while in the deterministic versions, the weights are
obtained by thresholding.

Let $W$ be a matrix or vector to be binarized. The stochastic BinaryConnect
update works as follows:
\begin{equation}
W_b=  \left \{
  \begin{array}{ccc}
  +1 & \text{with probability} & p=\sigma(W) \\
  -1 & \text{with probability} &1-p
  \end{array}
  \right.
\end{equation}
where $\sigma$ is the \textit{hard sigmoid} function:\\
\begin{equation}
\sigma(x)=\text{clip}\left ( \frac{x+1}{2},0,1 \right ) = \text{max} \left (0,\text{min}\left (1,\frac{x+1}{2}\right)\right)
\end{equation}
while in the deterministic BinaryConnect method, low-precision weights are obtained by thresholding the weight value by $0$.
\begin{equation}
W_b=  \left \{
  \begin{array}{ccc}
  +1 & \text{if} & W \geq 0 \\
  -1 & & \text{otherwise} 
  \end{array}
  \right.
\end{equation}

TernaryConnect allows weights to be additionally set to zero. Formally, the stochastic form
can be expressed as
\begin{equation}
W_{tern}=\text{sign}(W) \odot \text{binom}(p=\text{clip}(\text{abs}(2W),0,1))
\end{equation}
where $\odot $ is an element-wise multiplication. In the deterministic form, the weights are quantized depending on 2 thresholds:
\begin{equation}
W_{tern}=  \left \{
  \begin{array}{ccc}
  +1 & \text{if} & W > 0.5 \\
  -1 &  \text{if} &  W \leq -0.5\\
    0 & \text{otherwise} &
  \end{array}
  \right.
\end{equation}

Pow2-ternarization is another fixed-point oriented rounding method
introduced in \citep{Stromatias2015}. The precision of fixed-point numbers is
described by the Q\textit{m.f}notation, where \textit{m} denotes the number of
integer bits including the sign bit, and \textit{f} the number of fractional
bits. For example, Q1.1 allows $(-0.5, 0, 0.5)$ as values.
The rounding procedure works as follows:
We first round the values to be in range allowed by the number of integer bits:
\begin{equation}
W_{\text{clip}} =  \left \{
  \begin{array}{ccc}
  -2^m & \text{where} & W \leq -2^m \\
  W & \text{where} & -2^m < W < 2^m \\
  2^m & \text{where} & W \geq 2^m 
  \end{array}
  \right.
\end{equation}
We subsequently round the fractional part of the values:
\begin{equation}
W_{p2t}=\text{round}(2^f \cdot W_{\text{clip}})\cdot 2^{-f}
\end{equation}

\vspace*{-1mm}
\subsection{Exponential Quantization}   \label{quantization}
\vspace*{-1mm}
Quantizing the weight values to an integer power of 2 is also a way of storing
weights in low precision and eliminating multiplications. Since quantization
does not require a hard clipping of weight values, it scales well with weight
values.

Similar to the methods introduced above, we also have a deterministic and
stochastic way of quantization. For the stochastic quantization, we sample the
logarithm of weight value to be its nearest 2 integers, and the probability of
getting one of them is proportional to the distance of the weight value from
that integer. For weights with negative weight values, we take the logarithm of its absolute vale, but add their sign back after quantization. i.e.:
\begin{equation}
\label{quantize}
log_2{\left| W_b \right|}=  \left \{
  \begin{array}{ccc}
  \left\lceil log_{2}{\left| W \right|} \right\rceil & \text{with probability} & p=\frac{\left| W \right|}{2^{\left\lfloor log_{2}{\left| W \right|} \right\rfloor}}-1 \\
  \left\lfloor  log_{2}{\left| W \right|} \right\rfloor & \text{with probability} & 1-p
  \end{array}
  \right.
\end{equation}

For the deterministic version, we just set $log_2{W_b}=\left\lceil log_{2}{W} \right\rceil$ if the $p$ in Eq. \ref{quantize} is larger than 0.5. 

Note that we just need to store the logarithm of quantized weight values. The
actual instruction needed for multiplying a quantized number differs according
to the numerical format. For fixed point representation, multiplying by a
quantized value is equivalent to binary shifts, while for floating point
representation, that is equivalent to adding the quantized number's exponent to
the exponent of the floating point number. In either case, no complex operation
like multiplication would be needed.

\vspace*{-2mm}
\section{Low-Precision Recurrent Architectures}  \label{models}
\vspace*{-1mm}
\subsection{Vanilla Recurrent Networks}
\vspace*{-1mm}
As the most basic RNN structure, the vanilla RNN just adds a simple extension to
feed forward networks. Its hidden states updated from both the current input and the
 state at the previous time step: 
\begin{equation}
\label{rnn}
h_{t} = \sigma(W_{hh}h_{t-1} + W_{xh}x_{t} + b_h)
\end{equation}
where $\sigma(\dot)$ denotes the nonlinear activation function. The hidden
state can be followed by more layers to yield an output at each time step. For
example, in character-level language modeling, the output at each timestep is
set to be the probability of each character appearing at the next time step.
Thus there is a softmax layer that transforms the hidden state representation
into predictive probabilities.
\begin{equation}
\label{rnnoutput}
p_{t} = {\rm softmax}(W_{hx}h_t + b_x)
\end{equation}

In the low-precision version of the RNN, we just apply a quantization function
$\Omega(w)$ to each of the weights in the aforementioned RNN structure. Thus
all multiplications in the forward pass (except for the softmax normalization)
will be eliminated:
\begin{eqnarray}
h_{t} = \sigma(\Omega(W_{hh})h_{t-1} + \Omega(W_{xh})x_{t} + b_h) \\
p_{t} = {\rm softmax}(\Omega(W_{hx})h_t + b_x)
\end{eqnarray}
where $\Omega$ is applied element-wise to all weights in a given weight matrix.
We should note that, because of the quantization process, the derivative of the cost with
respect to weights is no longer smooth in the low-precision RNN (it is 0 almost
everywhere). We instead
compute the derivative with respect to the \emph{quantized} weights, and use that
derivative for weight update. In other words, the gradients are computed as if
the quantization operation were not there. This makes sense because we
can think of the quantization operation as {\em adding noise}:
\begin{equation}
   \Omega(w) = w + {\rm quantization\;\;''noise''}.
\end{equation}

\vspace*{-1mm}
\subsubsection{Hidden-state Stability}
\vspace*{-1mm}
We have observed among all the three RNN architectures
that BinaryConnect on the recurrent weights never worked.We should note that, the $\Omega()$ function in the recurrent direction has to
allow $0$ to be sampled.  We conjecture that this
is related to the stabilization of hidden states.

\begin{figure}
      \vspace*{-3mm}
    \centering
    \begin{subfigure}[!hb]{0.49\textwidth} 
    \includegraphics[width=1\textwidth]{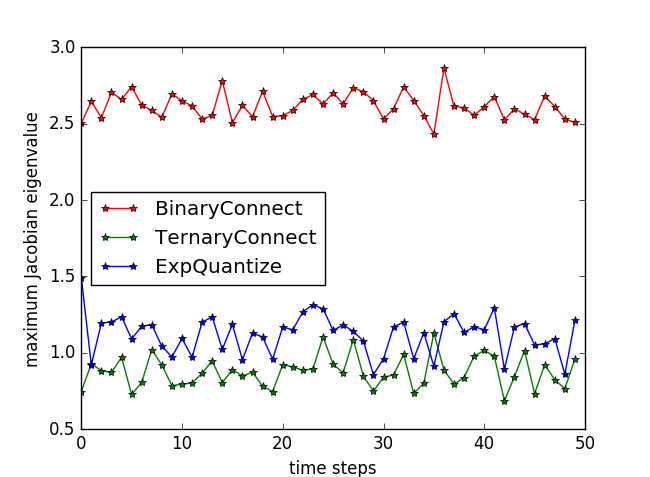}
    \label{fig:evalue}
    \caption{}\label{fig:hidstatea}
    \end{subfigure}
    ~
    \begin{subfigure}[!hb]{0.49\textwidth} 
    \includegraphics[width=1\textwidth]{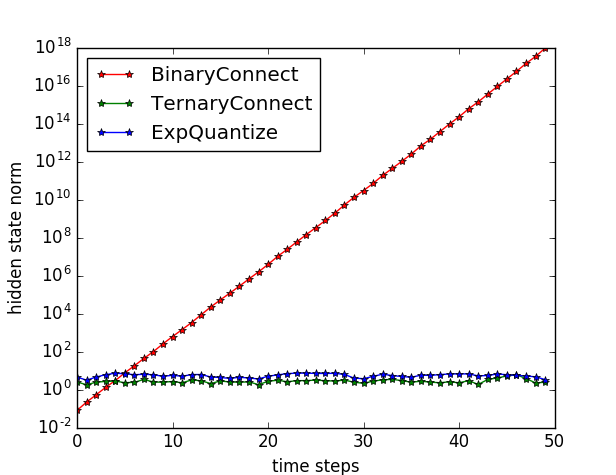}
	\label{fig:hidnorm}
    \caption{}\label{fig:hidstateb}
    \end{subfigure}
    \vspace*{-4mm}
    \caption{Hidden state statistics when BinaryConnect, TernaryConnect, and exponential quantization are applied. (a) Maximum eigenvalue of Jacobian matrix of the hidden-to-hidden connection, and (b) Hidden state norms, at each time step. The vertical axis is shown in log scale. BinaryConnect yields much larger derivatives, which probably explain the failure to train the RNNs properly. This makes the activations blow up exponentially over time when using ReLU as the activation function. (b)}
    \label{fig:hidstate}
\end{figure}

Consider the effect that BinaryConnect and TernaryConnect have on the
Jacobians of the state-to-state transition. In BinaryConnect, all entries in
$W_{hh}$ matrix are sampled to be $-1$ or $1$.
In LSTMs and GRUs, there is a strong near-1 diagonal in the
Jacobian because the gates are more often to be turned on, i.e., letting information
flow through it,
while the off-diagonal entries of the Jacobian tend to be much smaller
when the weights have not been quantized. 
However, when the true
value of a weight is near zero, its quantized value is stochastically sampled to
be $-1$ or $1$ with nearly equal probability. When near-0 off-diagonal
entries of a matrix of real values between -1 and 1 are randomly replaced by
values near +1 or -1, the magnitude of the weights increases and the
condition number of the matrix will tend to worsen
due to the presence of more near-0 eigenvalues. This could mean that
gradients tend to vanish faster, because a gradient vector $\frac{\partial C}{\partial h_t}$
could happen more often to have strong components in the directions of some of these
small eigenvectors. With larger eigenvalues of the Jacobian (observed, Fig.~\ref{fig:hidstate}(a)),
i.e. larger derivatives, we could also see gradients explode.

In Fig. \ref{fig:hidstatea}, where we use unbounded units (ReLU) as activation, 
if we look at the Jacobian of two neighboring hidden states ($\frac { \partial h_t }{
\partial h_{t-1} }$), we can see that the maximum eigenvalue of it is around
2.5 across all time steps, much larger than 1. As a consequence, hidden states
explode with respect to time steps, while this is not the case for
TernaryConnect and ExpQuantize. (Fig. \ref{fig:hidstateb})

On the other hand, if we allow $0$ (or a sufficiently small value) to be chosen in the sampling process, the effect of stochastic sampling on the Jacobians will not be that devastating. The Jacobian remains a quasi-diagonal matrix, which is well-conditioned.

In \citep{davidiclr} it was shown that
in a trained model, hidden state norms change in the first several timesteps,
but become stable afterwards. The model can work even better if  during
training we punish the changes of the norm of the hidden state from one time
step to the next.

\vspace*{-1mm}
\subsection{Long Short-Term Memory}
\vspace*{-1mm}
LSTMs \citep{Hochreiter1997} were first introduced in RNNs for sequence modeling. Its gate mechanism makes it a good option to deal with the vanishing gradient problem, since it can model long-term dependencies in the data.
To limit the numerical precision, we apply a rounding method $\Omega(w) $ to all or a subset of weights. 

\vspace*{-1mm}
\subsection{Gated Recurrent Unit}
\vspace*{-1mm}
GRUs \citep{Cho2014} can also be used in RNNs for modeling temporal sequences.

They involve less computation than the LSTM units, since they do not have an output gate, and are therefore sometimes preferred in large models. 
At timestep $t$, the state $s_t$ of a single GRU unit is computed as follows:
\begin{equation}
s_t = (1 - z) \odot h + z \odot s_{t-1}
\end{equation}
where $\odot$ denotes a element-wise multiplication. The update gate $z$ is computed with
\begin{equation}
z_t = \sigma (W_{xz} x_t + W_{hz} s_{t - 1} + b_z)
\label{formula:gru1}
\end{equation}
where $x_t$ is the input at timestep $t$, $W_{hz}$ is the
state-to-state recurrent weight matrix, $s_{t - 1}$ is the state at $t-1$, $W_{xz} $ is the input-to-hidden weight matrix, and $b_z$ is the bias.

The reset gate $r$ is computed as follows: 
\begin{equation}
r _t= \sigma (W_{xr} x_t  + W_{hr} s_{t - 1} + b_r)
\end{equation}
where\\
\begin{equation}
h _t= {tanh} (W_{xh} x_t  + W_{hh} (s_{t-1} \odot r)  + b_h)
\label{formula:gru2}
\end{equation}

In our experiments, the weights are rounded in the same way as the LSTMs. For example, for the $z$ gate, the input weight is rounded as follows: $z_t = \sigma (\Omega(W_{xz}) x_t + s_{t - 1} W_{hz} + b_z)$.

\vspace*{-2mm}
\section{Experimental Results and Discussion} \label{exp}
\vspace*{-1mm}

In the following experiments, we test the effectiveness of the different rounding methods on two different types of applications: character-level language modeling and speech recognition. The different RNN types (Vanilla RNN, GRU, and LSTM) are evaluated on experiments using four different datasets.

\vspace*{-1mm}
\subsection{Vanilla RNN}
\vspace*{-1mm}
We validate the low-precision vanilla RNN on 2 datasets: text8 and Penn Treebank Corpus (PTB).

The \textbf{text8} dataset contains the first 100M characters from Wikipedia, excluding
all punctuations. It does not discriminate between cases, so its alphabet has
only 27 different characters: the 26 English characters and space. We take the
first 90M characters as training set, and split them equally into sequences
with 50 character length each. The last 10M characters are split equally to
form validation and test sets. 

The \textbf{Penn Treebank Corpus} \citep{Taylor2003} contains 50 different characters,
including English characters, numbers, and punctuations. We follow the settings
in \citep{mikolov2012subword} to split our dataset, i.e., 5017k characters for
training set, 393k and 442k characters for validation and test set
respectively.

\paragraph{Model and Training}
The models are built to predict the next character given the previous ones, and
performances are evaluated with the bits-per-character (BPC) metric, which is $\log_2$ of
the perplexity, or the per-character log-likelihood (base 2).
We use a RNN with ReLU activation and 2048 hidden units. 
We initialize hidden-to-hidden weights as identity matrices, while
input-to-hidden and hidden-to-output matrices are initialized with uniform noise.

\begin{figure}
          \vspace*{-3mm}
    \centering
    \begin{subfigure}[b]{0.48\textwidth} 
    \includegraphics[width=1\textwidth]{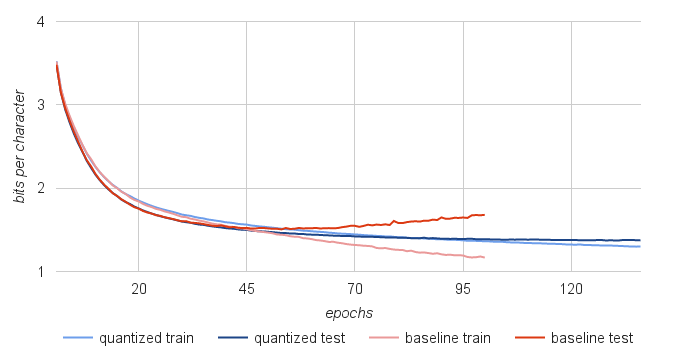}
    \label{fig:ptb}
    \caption{} \label{fig:ptbandtext8a}
    \end{subfigure}
    \begin{subfigure}[b]{0.48\textwidth} 
    \includegraphics[width=1\textwidth]{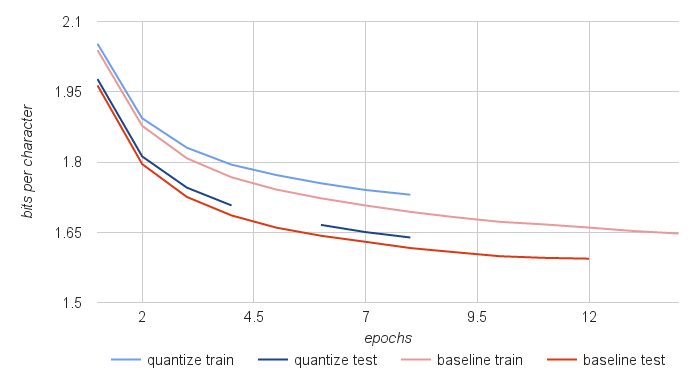}
	\label{fig:text8}
    \caption{} \label{fig:ptbandtext8b}
    \end{subfigure}
        \vspace*{-3mm}
    \caption{Training and test set performance of vanilla RNNs learning on PTB (a) and text8 (b) datasets.}
    \label{fig:ptbandtext8}
\end{figure}

We can see the regularization effect of stochastic quantization from the
results of the two datasets. 
In the PTB dataset, where the model size slightly overfits the dataset, the
low-precision model trained with stochastic quantization yields a test set
performance of 1.372 BPC, which surpasses its full precision baseline (1.505
BPC) by around 0.133 BPC (Fig.~\ref{fig:ptbandtext8}, left). From the figure we can see
that stochastic quantization does not significantly hurt training speed, and
manages to get better generalization when the baseline model begins to overfit.
On the other hand, we can also see from the results on the text8 dataset where
the same sized model now underfits, the low-precision model performs worse
(1.639 BPC) than its baseline (1.588 BPC). (Fig. ~\ref{fig:ptbandtext8b} and
Table \ref{table:roundingResults}).

\vspace*{-1mm}
\subsection{GRU RNNs}
\vspace*{-1mm}
This section presents results from the various methods to limit numerical precision in the weights and biases of GRU RNNs which are then tested on the TIDIGITS dataset.

\paragraph{Dataset}
TIDIGITS\citep{Leonard1993} is a speech dataset consisting of clean speech of spoken numbers from 326 speakers.
We only use single digit samples (zero to nine) in our experiments giving us 2464 training samples and 2486 validation samples. The labels for the spoken `zero' and `O' are combined into one label, hence we have 10 possible labels. We create MFCCs from the raw waveform and do leading zero padding to get samples of matrix size 39x200. The MFCC data is further whitened before use. We only use masking for processing the data with the RNN in some of the experiments.

\paragraph {Model and Training} The model has a 200 unit GRU layer followed by a 200 unit fully-connected ReLU layer. The output is a 10 unit softmax layer. Weights are initialized using the Glorot \& Bengio method \citep{Glorot2010}. The network is trained using Adam \citep{Kingma2014} and BatchNorm \citep{Ioffe2015} is not used. 
We train our model for up to 400 epochs with a patience setting of 100 (no early stopping before epoch 100). GRU results in Table \ref{table:roundingResults} are from 10 experiments, each experiment starts with a different random seed. This is done because previous experiments have shown that different random seeds can lead up to a few percent difference in final accuracy. We show average and maximum achieved performance on the validation set.

\paragraph{Binarization of Weights}
To evaluate weight binarization on a GRU, we trained our model with possible binary values \{-1,1\}, \{-0.5, 0.5\}, \{0, 1\}, \{-0.5, 0\} for the weights. Binarization was done only on the weights matrices $W_{xz}$, $W_{xr}$, $W_{xh}$. We ran each experiment once with stochastic binarization and once with deterministic binarization. As shown in Table\ref{table:roundingResults}, none of the combinations resulted in an increase in accuracy over chance after 400 training epochs. Also, doubling the number of GRU units to 400 did not help. We therefore concluded that GRUs do not function properly if all the weights are binarized. It has yet to be tested if at least a subset of the aforementioned weight matrices, or some of the hidden-to-hidden weight matrices could be binarized.

\paragraph{Effect of Pow2-Ternarization}
To assess the impact of weight ternarization, we trained our model and quantized the weights during training using pow2-ternarization with Q1.1. 

\begin{figure}
          \vspace*{-3mm}
    \centering
    \begin{subfigure}[b]{0.48\textwidth}        
\includegraphics[width=1\textwidth]{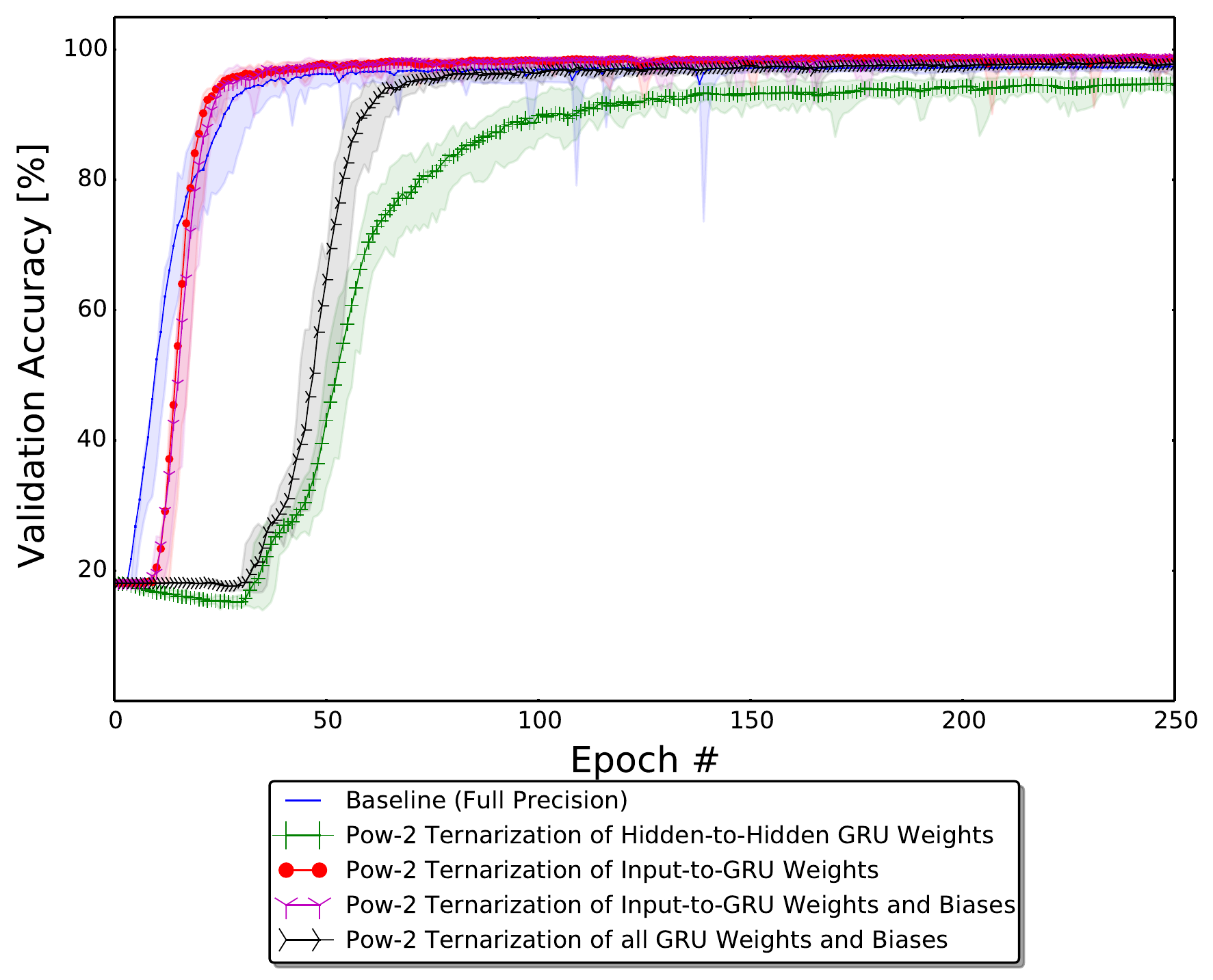}

\label{subfig:EffectDual-CopyRoundingConvergencePerformance_val_acc}
\caption{}

    \end{subfigure}
     ~
    \begin{subfigure}[b]{0.48\textwidth}
        \includegraphics[width=1\textwidth]{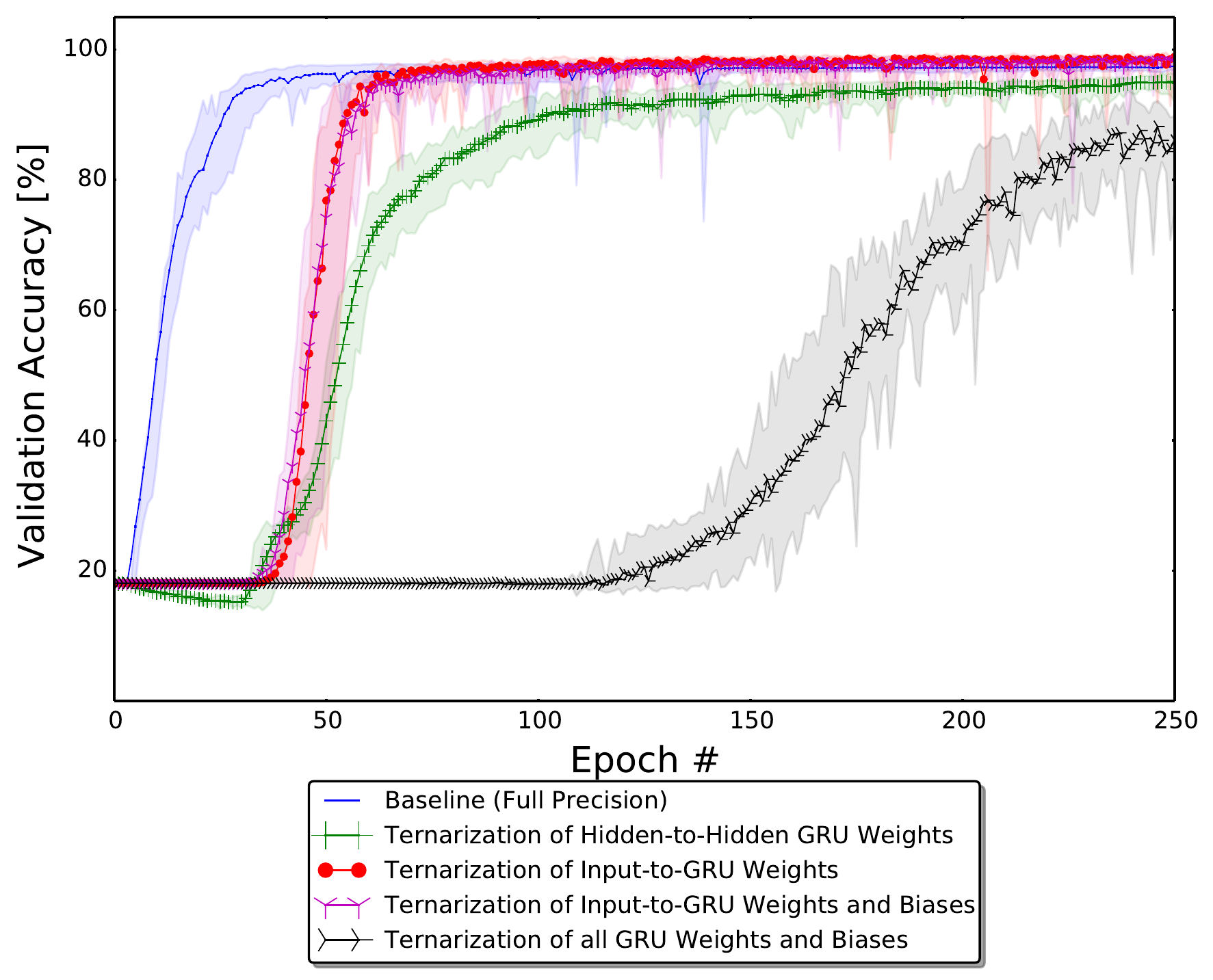}
     
        \label{fig:TernarizationGRUComponents_val_acc}
        \caption{}

    \end{subfigure}
           
    \caption{Effect of ternarization methods on GRUs. Thick curves show the mean of 10 runs, half-transparent areas around a curve show the variance. 
Accuracy from ternarization on hidden-to-GRU weights (green), input-to-GRU weights (red), input-to-GRU weights and biases (magenta), and on all GRU weights and biases (black) compared against baseline (blue).
(a) Pow2-ternarization. 
(b) Deterministic ternarization. Ternarization shows the same effects as pow2-ternarization, except for the case where all GRU weights and biases are ternarized (black). 
}
    \label{fig:GruComponents}
\end{figure}

Figure \ref{fig:GruComponents} (a)
shows how pow2-ternarization rounding applied on the different sets of GRU weights has an effect on convergence compared to the full-precision baseline. 
If full precision weights and biases are used, convergence starts after a few training epochs. As shown in Table \ref{table:roundingResults}, if pow2-ternarization is used on input-to-GRU weights, the top-1 improves to 99.3\%. Training takes 10 epochs longer before convergence starts, but then surpasses the baseline in terms of convergence speed, also the variance between the different runs is smaller compared to baseline runs. Limiting the precision of both input-to-GRU weights and biases leads to a similar training curve, but top-1score increases to 99.42\%. If pow2-ternarization is applied on all GRU weights (now also on $W_{hh}, W_{hz}, W_{hr}$ and biases, the top-1 decreases (though still higher than baseline) to to 99.1\%.\\
This shows that limiting the numerical precision of input-to-GRU weights and biases is beneficial for learning in this setup: although it slows down convergence, the final accuracy is higher than that of the baseline.
\begin{table}[ht]
\centering
\caption{Results from the three RNN types with the various strategies for limiting weight precision. GRU results are top-1, numbers in parenthesis are the average maximum accuracy. VRNN=Vanilla RNN. SB=Stochastic Binarization. DB=Deterministic Binarization.  ST=Stochastic Ternarization. DT=Deterministic Ternarization. PT=Pow2-Ternarization. EQ=Exponential Quantization.}
\label{table:roundingResults}
\resizebox{\textwidth}{!}{\begin{tabular}{l l l l l l l l l l }
\hline
\textbf{Dataset} & \textbf{RNN Type} & \textbf{Baseline} &\textbf{SB}&\textbf{DB}&\textbf{DT}& \textbf{ST}& \textbf{PT} & \textbf{EQ}  \\ \hline  \hline
text8 & VRNN & 1.588 BPC & N/A & N/A &  &  &  & 1.639 BPC   \\ \hline
PTC & VRNN & 1.505 BPC & N/A & N/A &  &  &  & 1.372 BPC  \\ \hline \hline
TIDIGITS & \begin{tabular}[c]{@{}l@{}}GRU \\ W\_x\end{tabular} & \begin{tabular}[c]{@{}l@{}}98.1\\ (97.52)\end{tabular} & 18.7 & 18.7 & \begin{tabular}[c]{@{}l@{}}99.67 \\ (99.44)\end{tabular} & \begin{tabular}[c]{@{}l@{}}98.23\\ (97.72)\end{tabular} & \begin{tabular}[c]{@{}l@{}}99.30\\ (99.05)\end{tabular} &  \begin{tabular}[c]{@{}l@{}}99.14\\ (98.73)\end{tabular} \\ \hline
 & \begin{tabular}[c]{@{}l@{}}GRU\\ W\_x, b\end{tabular} &  &  &  & \begin{tabular}[c]{@{}l@{}}99.55 \\ (99.24)\end{tabular} &  & \begin{tabular}[c]{@{}l@{}}99.42 \\ (99.06)\end{tabular} &  \begin{tabular}[c]{@{}l@{}}98.81\\ (98.25)\end{tabular}  \\ \hline
 & \begin{tabular}[c]{@{}l@{}}GRU\\ W\_h\end{tabular} &  &  &  & \begin{tabular}[c]{@{}l@{}}97.7 \\ (96.54)\end{tabular} &  & \begin{tabular}[c]{@{}l@{}}97.00 \\ (96.24)\end{tabular} &  \begin{tabular}[c]{@{}l@{}}98.68\\ (98.32)\end{tabular}  \\ \hline
 & \begin{tabular}[c]{@{}l@{}}GRU \\ all\end{tabular} &  &  &  & \begin{tabular}[c]{@{}l@{}}96.02 \\ (96.55)\end{tabular} &   \begin{tabular}[c]{@{}l@{}}96.55 \\ (96.02)\end{tabular} & \begin{tabular}[c]{@{}l@{}}99.18 \\ (98.86)\end{tabular} &\begin{tabular}[c]{@{}l@{}}99.10
\\ (98.76)\end{tabular}   \\ \hline \hline
WSJ & LSTM & \begin{tabular}[c]{@{}l@{}}11.16\\ WER \\ (Ep 60)\end{tabular} &  &  &  &  & \begin{tabular}[c]{@{}l@{}}10.49\\ WER \\ (Ep  87)\end{tabular} &    \\ \hline
\end{tabular}}
\end{table}

\paragraph{Effect of Ternarization}
Figure \ref{fig:GruComponents} (b) 
shows that we see the same effects as with pow2-ternarization, except for the case where we ternarize all weights and biases. With Pow-2 Ternarization we allow -0.5, 0, and 0.5 as values. With the default ternarization we allow -1, 0,1. This difference has a big impact on the hidden-to-hidden weight function, because if we apply ternarization there, we end up with lower-than-baseline performance and much slower convergence. On the other hand, if we apply ternarization only on the input-to-GRU weights, we get 99.67\%, the highest of all top-1 score of our TIDIGITS experiments. This leads us to conclude that different GRU components need different sets of allowed values to function in an optimal fashion. Indeed, if we change ternarization of all weights and biases to -0.5, 0, and 0.5 as allowed values, we see basically the same result as with pow2-ternarization.
Stochastic ternarization has not shown to be useful here. Convergence starts after 100 training epochs, and the average maximum and top-1 accuracy of 97.72\% and 98.23\% are almost at baseline level.

\vspace*{-1mm}
\subsection{LSTM RNNs}
\vspace*{-1mm}
Previous work had shown that some forms of network binarization work on small datasets but do not scale well to big datasets \citep{Rastegari2016}. To determine if low-precision networks still work on big datasets, we chose to train a large model on the WSJ dataset.

\paragraph{ Dataset}
The model is trained on the Wall Street Journal (WSJ) corpus (available at the LDC as LDC93S6B and LDC94S13B) where we use the 81 hour training set "si284".
The development set "dev93" is used for early stopping and the evaluation is performed on the test set "eval92". We use 40 dimensional filter bank features extended with deltas and delta-deltas, leading to 120 dimensional features per frame. Each dimension is normalized to have zero mean and unit variance over the training set. Following the text preprocessing in \citep{miao2015eesen}, we use 59 character labels for character-based acoustic modeling. Decoding with the language model is
performed on a recent proposed approach \citep{miao2015eesen} based on both Connectionist Temporal Classification (CTC) \citep{Graves2006} and weighted finite-state transducers (WFSTs) \citep{mohri2002weighted}.

\paragraph{Model and Training}
Both the limited precision model and the baseline model have 4 bidirectional LSTM layers with 250 units in both directions of each layer. In order to get the unsegmented character labels directly, we use CTC on top of the model. The baseline and the model are trained using Adam \citep{Kingma2014} with a fixed learning rate $1e^{-4}$. The weights are initialized following the scheme \citep{Glorot2010}. Notice that we do not regularize the model like injecting weight noise for simplicity, thus the
baseline results shown here could be worse than the recent published numbers on the same task \citep{graves2014towards, miao2015eesen}.  

\paragraph{Pow2-Ternarization on Weights}
The baseline achieves a word error rate (WER) of 11.16\% on the test set after training for 60 epochs, which took 8 days. The pow2-ternarization method has a considerably slower convergence similar to the GRU experiments. The model was trained for 3 weeks up to epoch 87, where it reaches an WER of 10.49\%.

\vspace*{-2mm}
\section{Conclusion and Outlook}
\vspace*{-1mm}
This paper shows for the first time how low-precision quantization of weights can be performed already during training effectively for RNNs.
We presented 3 existing methods and introduced 1 new method of limiting the numerical precision.  We used the different methods on 3 major RNN types and determined how the limited numerical precision affects network performance across 4 datasets.

In the language modeling task, the low-precision model surpasses its full-precision baseline by a large gap (0.133 BPC) on the PTB dataset. We also show that the model will work better if put in a slightly overfitting setting, so that the regularization effect of stochastic quantization will begin to function. In the speech recognition task, we show that it is not possible to binarize weights of GRUs while maintaining their functionality. We conjecture that the better performance from ternarization is due to a reduced variance of the weighted sums (when a near-zero real value is quantized to +1 or -1, this introduces substantial variance), which could be more harmful in RNNs because the same weight matrices are used over and over again along the temporal sequence.
Furthermore, we show that weight and bias quantization methods using ternarization, pow2-ternarization, and exponential quantization, can improve performance over the baseline on the TIDIGITs dataset. The successful outcome of these experiments means that lower resource requirements are needed for custom implementations of RNN models.

\section{Acknowledgments}
We are grateful to INI members Danny Neil, Stefan Braun, and Enea Ceolini, and MILA members Philemon Brakel, Mohammad Pezeshki, and Matthieu Courbariaux, for useful discussions and help with data preparation. 
We thank the developers of Theano\citep{2016arXiv160502688short}, Lasagne, Keras, Blocks\citep{VanMerrienboer2015}, and Kaldi\citep{Povey_ASRU2011}.\\
The authors acknowledge partial funding from the Samsung Advanced Institute of Technology, University of Zurich, NSERC,
CIFAR and Canada Research Chairs.\\

\small
\FloatBarrier

\bibliographystyle{plainnat}
\bibliography{ott_lin_biblio}
\end{document}